\documentclass{bmvc2k}

\usepackage{amsmath}
\usepackage{amssymb}
\usepackage{xcolor}
\usepackage{multirow}
\usepackage{booktabs}

\title{RCFormer: Reliability-aware Context Transformer for Occlusion-Robust Facial Landmark Detection}

\addauthor{Ashish Kumar$^{*}$}{ashish_k@cs.iitr.ac.in}{1}
\addauthor{Prashant Kumar$^{*}$}{prashant.kumar@afinit.com}{2}
\addauthor{Mallika Garg$^{*}$}{mallika@ec.iitr.ac.in}{3}

\addinstitution{
Indian Institute of Technology Roorkee \\ Roorkee, INDIA
}
\addinstitution{
Afinit Co. Ltd\\
Seoul, Republic of Korea
}
\addinstitution{
Indian Institute of Technology Kharagpur\\
Kharagpur, INDIA
}

\runninghead{Kumar, Kumar, Garg}{RCFormer}

\begin{document}

\maketitle

\begingroup
\renewcommand{\thefootnote}{\fnsymbol{footnote}}
\footnotetext[1]{All authors contributed equally to this work.}
\endgroup

\begin{abstract}

Facial landmark detection has advanced substantially under regular imaging conditions, yet occlusion, poor illumination, and large pose changes remain major sources of error because local facial evidence can become incomplete or misleading. Context-guided latent patch recovery offers a promising way to address this issue by inferring unreliable patch representations from surrounding facial context before heatmap prediction. However, the recovered context can itself be unreliable when it is weakly adapted to the input face, influenced by corrupted neighboring patches, or blended without preserving local spatial coherence. We propose RCFormer, a Reliability-aware Context Transformer for occlusion-robust facial landmark detection. RCFormer strengthens latent patch recovery with three reliability-oriented components: visible-context conditioning (VCC), which adapts the context representation using a visibility-weighted global facial summary; visibility-guided context aggregation (VGCA), which reduces the influence of unreliable source patches during attention; and spatial residual correction (SRC), which refines blended latent features using neighboring patch information. The recovered heatmaps are then used as reliability-aware structural cues for existing landmark detection networks. Experiments show that RCFormer improves heatmap recovery and achieves strong landmark localization results on the WFLW, COFW, and 300W datasets. Code is available at: \url{https://github.com/havelhakimi/RCFormer}
\end{abstract}

\section{Introduction}

Facial landmark detection (FLD) aims to identify semantically meaningful facial keypoints, such as those around the eyes, nose, mouth, and facial contour. It serves as a fundamental component in many face-related applications, including face recognition~\cite{parkhi2015deep}, facial expression analysis~\cite{barsoum2016training}, head-pose estimation~\cite{hempel2022}, and augmented reality~\cite{kang2021}. With advances in deep neural networks, FLD methods have achieved strong performance under regular imaging conditions. However, their performance often degrades in challenging scenarios involving occlusion, extreme illumination, and large head-pose variation, where local facial evidence may be incomplete or misleading. In such cases, features extracted from affected regions can become corrupted, making accurate landmark localization difficult. Therefore, improving FLD robustness requires not only detecting reliable facial evidence but also recovering useful structural cues when local observations are degraded or unavailable.

\par A promising direction in occlusion-robust FLD is context-guided latent patch recovery, which addresses unreliable facial regions in the latent feature space before landmark prediction. Rather than relying only on features extracted from a potentially occluded patch, this approach estimates its representation from surrounding facial context. This is motivated by the strong structural dependency among facial regions: visible areas around the eyes, nose, mouth, and contour can provide cues about the expected representation of a missing or corrupted region. The recovered patch feature therefore serves as a context-derived alternative when the target patch is unreliable or unavailable. By recovering degraded latent features before heatmap decoding, context-guided recovery can generate more reliable heatmaps and improve landmark localization under partial visibility. ORFormer~\cite{orformer} is a representative method in this direction, introducing messenger tokens that aggregate information from non-target patches and use it for occlusion-aware feature recovery.

\par Although context-guided latent patch recovery offers an effective way to handle corrupted facial regions, it introduces a second reliability problem: the recovered context is useful only when the context itself is trustworthy. A context query that is mainly shaped by learned tokens may not fully adapt to the visible structure of the current face, where pose, expression, identity, and occlusion patterns can vary significantly. Moreover, the surrounding context may also contain unreliable patches, so aggregating features without explicitly suppressing corrupted source regions can propagate degraded information into the recovered representation. Finally, recovering each latent patch through patch-wise blending can overlook the local continuity of facial contours and neighboring structures, leading to features that are context-derived but not spatially coherent. These limitations motivate a recovery mechanism that is image-adaptive, visibility-aware, and spatially refined before heatmap decoding.

\par Building on context-guided latent patch recovery, we introduce RCFormer, a Reliability-aware Context Transformer for occlusion-robust FLD. RCFormer addresses the reliability limitations above by making context recovery image-adaptive, visibility-aware, and spatially coherent. First, visible-context conditioning (VCC) adapts the context representation using a visibility-weighted global facial summary, allowing the recovery branch to reflect the visible structure of the current face. Second, visibility-guided context aggregation (VGCA) suppresses unreliable source patches during attention, so corrupted regions contribute less to the recovered representation. Third, spatial residual correction (SRC) refines the blended latent features with neighboring patch information, improving local continuity especially in regions predicted to be unreliable.

\par We incorporate the recovered heatmaps generated by RCFormer as reliability-aware structural cues for existing landmark detection networks. These heatmaps transfer recovered facial structure to the landmark network and improve prediction under partial visibility. Experiments on WFLW~\cite{wu2018wflw}, COFW~ \cite{burgos2013cofw}, and 300W~\cite{sagonas2013300w} show that RCFormer improves heatmap quality and achieves strong landmark localization performance. Our main contributions are summarized as follows:
\begin{itemize}
    \item We identify a reliability gap in context-guided latent patch recovery, where the recovered context may be insufficiently adaptive to the input face, affected by unreliable source patches, or weakly constrained by local spatial structure.
    \item We propose RCFormer, a reliability-aware context transformer with visible-context conditioning (VCC), visibility-guided context aggregation (VGCA), and spatial residual correction (SRC) to improve context recovery.
    \item We validate RCFormer on WFLW, COFW, and 300W, showing improved heatmap recovery and strong landmark localization under challenging facial conditions.
\end{itemize}

\section{Related Work}
\paragraph{Facial Landmark Detection.}
Existing FLD methods mainly follow coordinate-regression or heatmap-regression formulations. Coordinate-regression methods directly predict landmark locations from CNN features using regression heads~\cite{feng2018wing,guo2019pfld}. Feng et al.~\cite{feng2018wing} introduce wing loss for improved landmark supervision, Wu et al.~\cite{wu2018wflw} use facial boundary information to impose structural constraints, and Miao et al.~\cite{miao2018} model complex facial shape variations through Fourier feature pooling. 
Transformer-based methods further use attention and structured representation learning for relational modeling~\cite{lha23}: Xia et al.~\cite{xia2022slpt} propose a coarse-to-fine decoder over sparse local patches, while Li et al.~\cite{li2022repformer} learn landmark queries from pyramid features. Heatmap-regression methods predict spatial response maps for landmarks, with Newell et al.~\cite{newell2016stacked} using stacked hourglass networks and later works improving heatmap supervision and localization accuracy~\cite{wang2019awing,huang2021adnet,zhou2023star}. To reduce reliance on non-differentiable heatmap decoding, Jin et al.~\cite{jin2021pipnet} reformulate prediction as confidence and offset estimation, Huang et al.~\cite{huang2021adnet} and Zhou et al.~\cite{zhou2023star} design losses using both heatmap and landmark supervision, and Micaelli et al.~\cite{micaelli2023ldeq} use a deep equilibrium formulation for cascaded landmark refinement. This ability to preserve facial structure while supporting both heatmap and landmark supervision has made heatmap-based methods highly effective in recent FLD, although their predictions can still degrade under occlusion.

\paragraph{Occlusion-aware Facial Landmark Detection and Feature Recovery.}

Several FLD methods improve robustness by modeling occlusion, visibility, or uncertainty during landmark prediction. Kumar et al.~\cite{kumar2020luvli} estimate landmark location with uncertainty and visibility likelihood, while Li et al.~\cite{li2024cascaded} jointly predict landmarks, occlusion probability, and head pose. Wu et al.~\cite{wu2015robust} also address landmark localization under large pose and occlusion. While effective, these methods often require visibility or occlusion-related supervision. Other works use local-predictor agreement or facial context to handle missing evidence: Burgos-Artizzu et al.~\cite{burgos2013cofw} identify reliable regions through consensus among patch-based regressors, Merget et al.~\cite{merget2018robust} incorporate global context in a fully convolutional framework, Zhu et al.~\cite{zhu2019occlusion} model geometric relationships between facial components, and Zhu et al.~\cite{zhu2022occlusion} capture hierarchical facial-component relations. Although these approaches improve robustness under partial visibility, they mainly down-weight, bypass, or compensate for unreliable regions.

A more direct alternative is to recover the latent representation of corrupted regions from visible facial context before heatmap decoding. Transformers provide a natural tool for such recovery because they model long-range token interactions, but their attention can become unreliable when some tokens are corrupted or occluded. Several methods recover degraded visual representations in related tasks, including occluded person re-identification~\cite{xu2022learning}, hand mesh estimation~\cite{park2022handoccnet}, blind face restoration~\cite{codeformer}, and shadow-robust image generation~\cite{lumi26}. However, these methods are designed for different settings, such as multi-frame recovery or generic face restoration, and are not directly tailored to single-image FLD heatmap recovery. For single-image FLD, this recovery problem is more specifically addressed by context-guided latent patch recovery, where corrupted facial patch representations are estimated from surrounding facial context before heatmap decoding. ORFormer~\cite{orformer} follows this idea by introducing messenger tokens that exclude the target patch during context aggregation and provide context-derived features for occlusion-aware blending with regular latent features. However, its recovery quality still depends on how reliably the context is formed: the context representation is weakly adapted to the current face, unreliable surrounding patches may still affect aggregation, and patch-wise blending may not fully preserve local spatial continuity. RCFormer addresses these limitations through visible-context conditioning (VCC), visibility-guided context aggregation (VGCA), and spatial residual correction (SRC).

\begin{figure}[t]
    \centering
    \includegraphics[width=\linewidth]{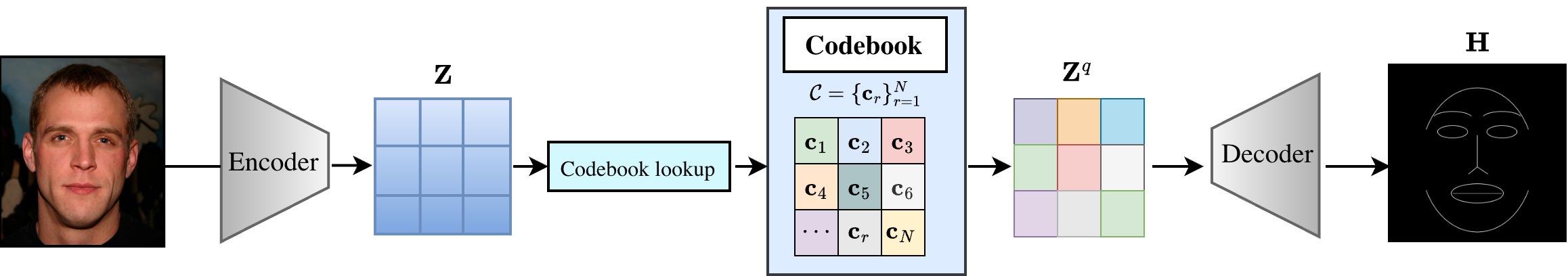}
    \caption{Overview of the quantized heatmap prior learning stage. A clean face image is encoded into latent features, quantized through a learned codebook, and decoded into an edge heatmap representation.}
    \label{fig:qhp}
\end{figure}

\section{Method}
\label{sec:method}

\begin{figure*}[t]
    \centering
    \includegraphics[width=\textwidth]{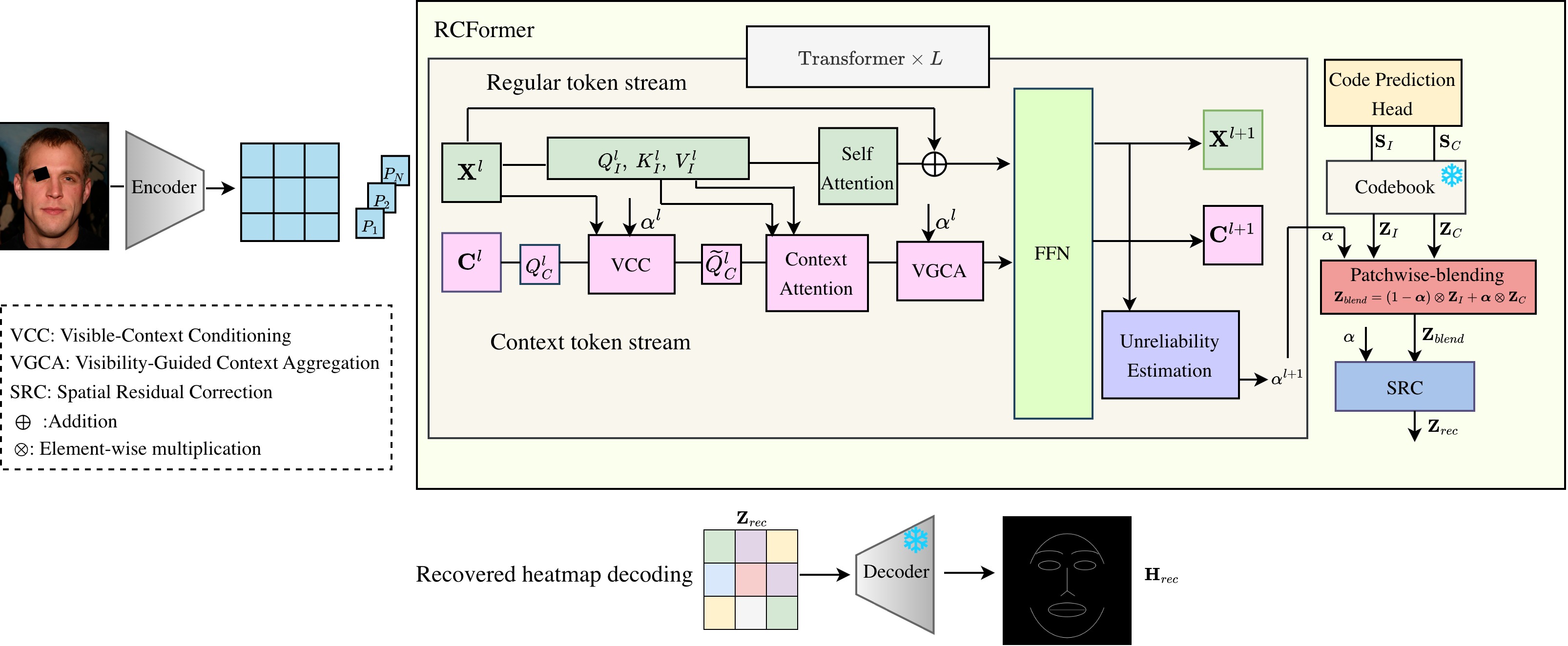}
\caption{Architecture of RCFormer for reliable context-guided latent recovery. Given latent patches from an occluded input, RCFormer maintains regular and context token streams within an $L$-layer reliability-aware recovery transformer. Visible-context conditioning (VCC) conditions context queries using reliable global evidence, while visibility-guided context aggregation (VGCA) suppresses unreliable source patches during context aggregation. The final regular and context tokens predict code sequences $\mathbf{S}_{I}$ and $\mathbf{S}_{C}$, which are mapped through the fixed codebook to obtain $\mathbf{Z}_{I}$ and $\mathbf{Z}_{C}$. The final unreliability map $\boldsymbol{\alpha}$ guides patch-wise blending and gates spatial residual correction (SRC) to produce $\mathbf{Z}_{rec}$. The fixed decoder maps $\mathbf{Z}_{rec}$ to the recovered heatmap $\mathbf{H}_{rec}$. For downstream FLD, $\mathbf{H}_{rec}$ is concatenated with an early-stage detector feature map and fused using a lightweight $1\times1$ convolutional block before landmark prediction.}
    \label{fig:rcformer_arch}
\end{figure*}

This section presents RCFormer for reliability-aware context-guided latent recovery. Following codebook-based recovery works~\cite{codeformer,orformer}, we first adopt vector quantization~\cite{vqvae} to learn a quantized heatmap prior (Section~\ref{sec:qhp}; Figure~\ref{fig:qhp}). After pre-training, the learned codebook and decoder are fixed and used as the prior for latent recovery. We then introduce RCFormer, which recovers reliable latent features from partially visible facial inputs using visible-context conditioning, visibility-guided context aggregation, and spatial residual correction (Section~\ref{rcformer}; Figure~\ref{fig:rcformer_arch}). Finally, the recovered heatmaps are used as structural cues for downstream FLD networks (Section~\ref{fldusage}).

\subsection{Learning the Quantized Heatmap Prior}
\label{sec:qhp}

Following codebook-based recovery works~\cite{codeformer,orformer}, we first learn a quantized heatmap prior before training RCFormer. This prior is trained on clean facial structures so that the learned codebook and decoder capture heatmap patterns under reliable facial evidence. Such a clean heatmap-aligned prior reduces the uncertainty of later recovery, where corrupted latent patches must be mapped back to plausible facial structures. Specifically, we train a VQVAE-style~\cite{vqvae} heatmap generator that quantizes encoder features through a learned codebook and decodes the quantized representation into facial edge heatmaps. We follow the edge heatmap definition used in FLD works~\cite{wu2018wflw,orformer}, where facial contours are represented as structural heatmap channels. Heatmap generation details and dataset-specific landmark-to-edge mappings are provided in the supplementary material. After training, the codebook and decoder are fixed and used as the prior for the subsequent RCFormer recovery stage. Figure~\ref{fig:qhp} illustrates this prior learning stage.

A clean facial image $\mathbf{I}\in\mathbb{R}^{h\times w\times 3}$ is first mapped by an encoder $E$ to a latent representation $\mathbf{Z}=E(\mathbf{I})$, where $\mathbf{Z}\in\mathbb{R}^{m\times n\times d}$. We denote the $K=mn$ latent tokens in $\mathbf{Z}$ as $\{\mathbf{z}_k\}_{k=1}^{K}$, with each token $\mathbf{z}_k\in\mathbb{R}^{d}$. Given a learnable codebook $\mathcal{C}=\{\mathbf{c}_r\}_{r=1}^{N}$, where each code vector $\mathbf{c}_r\in\mathbb{R}^{d}$, each latent token is quantized by assigning it to its nearest codebook entry:
\begin{equation}
\label{eq:quantized_prior_assignment}
a_k=\arg\min_{r\in\{1,\ldots,N\}}\|\mathbf{z}_k-\mathbf{c}_r\|_2^2,
\qquad
\mathbf{z}^{q}_k=\mathbf{c}_{a_k}.
\end{equation}
The resulting index sequence is $\mathbf{a}=\{a_k\}_{k=1}^{K}\in\{1,\ldots,N\}^{K}$, and the quantized latent representation is $\mathbf{Z}^{q}=\{\mathbf{z}^{q}_k\}_{k=1}^{K}\in\mathbb{R}^{m\times n\times d}$. The decoder $D$ maps this quantized representation to predicted edge heatmaps $\mathbf{H}=D(\mathbf{Z}^{q})$, where $\mathbf{H}\in\mathbb{R}^{h\times w\times N_E}$ and $N_E$ denotes the number of facial edge heatmap channels.

The generator is trained with heatmap supervision and vector-quantization regularization. The heatmap term penalizes the difference between the predicted edge heatmaps $\mathbf{H}$ and the target heatmaps $\hat{\mathbf{H}}$, while the VQ term aligns the encoder features with their selected codebook vectors:
\begin{equation}
\label{eq:qhg_indiv_losses}
\mathcal{L}_{hm} = \|\mathbf{H}-\hat{\mathbf{H}}\|_2^2,
\quad
\mathcal{L}_{vq}
=
\|\operatorname{sg}(\mathbf{Z})-\mathbf{Z}^{q}\|_2^2
+
\beta\|\mathbf{Z}-\operatorname{sg}(\mathbf{Z}^{q})\|_2^2 .
\end{equation}
Here, $\operatorname{sg}(\cdot)$ denotes the stop-gradient operation, and $\beta$ controls the commitment term. The overall objective for training the quantized heatmap generator is
\begin{equation}
\label{eq:qhg_objective}
\mathcal{L}_{QHP}
=
\mathcal{L}_{hm}
+
\lambda_{vq}\mathcal{L}_{vq},
\end{equation}
where $\lambda_{vq}$ balances heatmap reconstruction and vector-quantization regularization. After training, the codebook and decoder are kept fixed for the subsequent RCFormer recovery stage.

\subsection{RCFormer for Reliable Context-Guided Latent Recovery}
\label{rcformer}

For faces affected by occlusion or other visibility degradations, encoder latent patches can become inconsistent with the clean facial structures captured by the learned heatmap prior. Building on context-guided latent patch recovery, RCFormer predicts reliable latent codes before heatmap decoding. However, context-guided recovery can still fail when the context branch copies from the target patch, attends to corrupted surrounding patches, or produces features that are not locally coherent.

To address these issues, RCFormer combines an $L$-layer reliability-aware token recovery module with code prediction, latent blending, and spatial residual correction. It introduces three reliability-oriented components: visible-context conditioning (VCC), visibility-guided context aggregation (VGCA), and spatial residual correction (SRC). Given latent patches from the encoder, RCFormer predicts a patch-wise unreliability map $\boldsymbol{\alpha}$, a regular code sequence $\mathbf{S}_I$, and a context code sequence $\mathbf{S}_C$. The predicted codes are mapped to quantized latent features using the fixed codebook, combined according to $\boldsymbol{\alpha}$, and refined to obtain $\mathbf{Z}_{rec}$. The fixed decoder then maps $\mathbf{Z}_{rec}$ to the recovered heatmaps $\mathbf{H}_{rec}$. Figure~\ref{fig:rcformer_arch} illustrates the overall RCFormer architecture, including the regular and context token streams, VCC, VGCA, code prediction, latent blending, and SRC.

\paragraph{Regular and Context Token Encoding.}
RCFormer maintains two token streams over the encoder latent patches. At each layer, the regular stream updates the observed patch tokens $\mathbf{X}^{l}$ using standard self-attention followed by an FFN, with projections $(\mathbf{Q}^{l}_{I},\mathbf{K}^{l}_{I},\mathbf{V}^{l}_{I})$ computed from $\mathbf{X}^{l}$. These tokens retain the available image evidence for code prediction. In parallel, the context stream maintains one context token for each patch, denoted by $\mathbf{C}^{l}$. Each context token attends to the regular patch tokens, using its own query while taking $\mathbf{K}^{l}_{I}$ and $\mathbf{V}^{l}_{I}$ from the regular stream as keys and values to form an alternative representation of the corresponding patch from surrounding facial regions. This provides the base context stream, which is further conditioned by VCC and visibility-reweighted by VGCA in the following steps.

To prevent the context representation from directly copying its own patch, RCFormer applies a hard self-exclusion mask in the base context attention. For the $i$-th context token and the $j$-th regular token, the attention logit is defined as
\begin{equation}
\label{eq:self_excluded_context_attention}
\mathbf{B}^{l}_{ij}
=
\frac{\mathbf{q}^{l}_{C,i}(\mathbf{k}^{l}_{I,j})^{\top}}{\sqrt{d}}
\quad \text{for } i\neq j,
\qquad
\mathbf{B}^{l}_{ii}=-\infty ,
\end{equation}
where $\mathbf{q}^{l}_{C,i}$ is the query of the $i$-th context token and $\mathbf{k}^{l}_{I,j}$ is the key of the $j$-th regular token. The context tokens are then updated by multiplying $\operatorname{softmax}(\mathbf{B}^{l})$ with the regular-token value matrix $\mathbf{V}^{l}_{I}$. This gives each patch a context-derived counterpart inferred from other facial regions, while keeping direct self-patch information blocked.


\paragraph{Unreliability Estimation.}
RCFormer estimates patch unreliability from the discrepancy between each regular token and its context-derived counterpart. For the $k$-th patch after layer $l$, we compute
\begin{equation}
\label{eq:unreliability_estimation}
\alpha^{l+1}_{k}
=
\sigma\!\left(
\mathbf{W}^{l+1}_{\alpha}
\operatorname{dist}(\mathbf{x}^{l+1}_{k},\mathbf{c}^{l+1}_{k})
\right),
\end{equation}
where $\operatorname{dist}(\cdot,\cdot)$ is the element-wise squared difference, $\mathbf{W}^{l+1}_{\alpha}$ maps the discrepancy vector to a scalar, and $\sigma(\cdot)$ maps $\alpha^{l+1}_{k}$ to $(0,1)$. A larger value indicates greater unreliability. We use this discrepancy because it directly measures disagreement between the observed patch representation and its context-derived estimate. In our analysis, attention-entropy-based unreliability and a $3\times3$ convolutional head applied to the regular--context discrepancy produced higher WFLW heatmap L2 loss. Moreover, $\alpha$ acts as a soft weight rather than a hard mask, so inaccurate estimates modulate rather than discard patch information. We set $\boldsymbol{\alpha}^{1}=\mathbf{0}$ and denote the final-layer estimate by $\boldsymbol{\alpha}$.

\paragraph{Visible-Context Conditioning (VCC).}
A context token initialized only from a learned query is weakly image-adaptive, since its context formation is not explicitly conditioned on the visible structure of the current face. To make the context branch input-adaptive before aggregation, RCFormer conditions each context query on a visibility-weighted global summary of reliable facial evidence. Given the unreliability map from the previous layer, we first compute visibility weights:
\begin{equation}
\label{eq:vcc_weights}
w^{l}_{j}
=
\frac{1-\operatorname{sg}(\alpha^{l}_{j})}
{\sum_{t=1}^{K}\left(1-\operatorname{sg}(\alpha^{l}_{t})\right)+\epsilon}.
\end{equation}
Using these weights, the global visible context is obtained as
\begin{equation}
\label{eq:vcc_global_context}
\mathbf{g}^{l}
=
\sum_{j=1}^{K} w^{l}_{j}\mathbf{x}^{l}_{j} .
\end{equation}
The context query is then conditioned by adding a projected version of this global visible context:
\begin{equation}
\label{eq:vcc_conditioned_query}
\tilde{\mathbf{q}}^{l}_{C,i}
=
\mathbf{q}^{l}_{C,i}
+
\mathbf{W}^{l}_{g}\mathbf{g}^{l}.
\end{equation}
Here, $\mathbf{W}_g^l$ is a learnable projection that maps $\mathbf{g}^l$ to the context-query dimension. The operator $\operatorname{sg}(\cdot)$ stops gradients through the unreliability weights, so $1-\boldsymbol{\alpha}^{l}$ is used as a visibility estimate rather than a shortcut for optimizing query conditioning. At the first layer, where $\boldsymbol{\alpha}^{1}=\mathbf{0}$, the summary reduces to an average facial context; in later layers, it emphasizes patches predicted to be reliable. The conditioned query $\tilde{\mathbf{q}}^{l}_{C,i}$ replaces the base query $\mathbf{q}^{l}_{C,i}$ in Eq.~\ref{eq:self_excluded_context_attention} when computing the context-attention score matrix.



\paragraph{Visibility-Guided Context Aggregation (VGCA).}
Using the VCC-conditioned queries, RCFormer computes self-excluded context attention and then modulates the resulting attention probabilities according to source reliability. Let $\mathbf{A}^{l}=\operatorname{softmax}(\mathbf{B}^{l})$ denote the context-attention matrix obtained after VCC-based query conditioning and hard self-exclusion. Since some source patches may themselves be corrupted, we use $1-\boldsymbol{\alpha}^{l}$ as the source visibility estimate:
\begin{equation}
\label{eq:visibility_guided_attention}
\tilde{\mathbf{A}}^{l}_{ij}
=
\frac{
\mathbf{A}^{l}_{ij}(1-\alpha^{l}_{j})
}{
\sum_{t=1}^{K}\mathbf{A}^{l}_{it}(1-\alpha^{l}_{t})+\epsilon
}.
\end{equation}

We apply visibility guidance after softmax so that attention first captures query--key compatibility, after which $(1-\alpha^{l}_{j})$ reweights the normalized probabilities according to source reliability. This separates semantic matching from reliability modulation without altering the raw attention-logit scale. The subsequent renormalization suppresses unreliable source patches while increasing the relative contribution of reliable ones. The context tokens are then updated as $\mathbf{C}^{l+1}=\operatorname{FFN}(\tilde{\mathbf{A}}^{l}\mathbf{V}^{l}_{I})$.

\paragraph{Code Prediction and Spatial Residual Correction (SRC).}
After the final layer of the reliability-aware token recovery module, the regular tokens $\mathbf{X}^{L+1}$ and the context tokens $\mathbf{C}^{L+1}$ are used to predict two code sequences, denoted by $\mathbf{S}_{I}$ and $\mathbf{S}_{C}$, respectively. The predicted codes are mapped through the fixed codebook $\mathcal{C}$ to obtain the quantized latent features $\mathbf{Z}_{I}$ and $\mathbf{Z}_{C}$. Here, $\mathbf{Z}_{I}$ preserves information from the regular patch stream, while $\mathbf{Z}_{C}$ provides a context-derived alternative inferred from surrounding facial regions.

Using the final unreliability map $\boldsymbol{\alpha}$, RCFormer first forms a patch-wise blended latent representation:
\begin{equation}
\label{eq:rcformer_blend}
\mathbf{Z}_{blend}
=
(1-\boldsymbol{\alpha})\otimes \mathbf{Z}_{I}
+
\boldsymbol{\alpha}\otimes \mathbf{Z}_{C},
\end{equation}
where $\otimes$ denotes element-wise multiplication between a patch-wise weight map and latent features along the channel dimension. This blend trusts the regular feature on reliable patches and shifts toward the context-derived feature on unreliable patches.

Patch-wise blending treats each latent position independently and may not fully preserve the local continuity of facial structures. RCFormer therefore applies spatial residual correction to refine the blended latent feature with neighborhood-aware information. A lightweight depthwise spatial operator extracts local context from $\mathbf{Z}_{blend}$, and a linear projection layer predicts the residual correction from both the blended feature and its local response:
\begin{equation}
\label{eq:src_recovery}
\mathbf{Z}_{rec}
=
\mathbf{Z}_{blend}
+
\boldsymbol{\alpha}\otimes
\phi\!\left(
\left[
\mathbf{Z}_{blend},
\psi_{sp}(\mathbf{Z}_{blend})
\right]
\right),
\end{equation}
where $\psi_{sp}(\cdot)$ denotes the depthwise spatial operator, $\phi(\cdot)$ is a linear projection layer that predicts the residual correction, and $[\cdot,\cdot]$ denotes channel-wise concatenation. Since the correction is weighted by $\boldsymbol{\alpha}$, it is emphasized on unreliable regions while reliable patches remain largely preserved. Finally, the fixed decoder maps $\mathbf{Z}_{rec}$ to the recovered heatmaps $\mathbf{H}_{rec}$.


\paragraph{Training Objective.}
After learning the quantized heatmap prior, we train RCFormer while keeping the codebook $\mathcal{C}$ and decoder $D$ fixed, and fine-tune the encoder together with RCFormer. Although encoder adaptation may shift its feature distribution, the fixed codebook and decoder, together with code-prediction and heatmap-reconstruction supervision, constrain the recovered representation toward the learned clean heatmap prior. For supervision, each training image is passed through the learned quantized heatmap prior to obtain the target code sequence $\mathbf{a}$ by the nearest-code assignment in Eq.~\ref{eq:quantized_prior_assignment}. This target sequence supervises both predicted code sequences $\mathbf{S}_{I}$ and $\mathbf{S}_{C}$ using cross-entropy over code indices. We also apply a heatmap reconstruction loss,
$\mathcal{L}_{hm}=\|\mathbf{H}_{rec}-\hat{\mathbf{H}}\|_2^2$,
between the recovered heatmaps $\mathbf{H}_{rec}$ and the target heatmaps $\hat{\mathbf{H}}$. The complete objective is
\begin{equation}
\label{eq:rcformer_objective}
\mathcal{L}_{RCFormer}
=
\mathcal{L}_{code}(\mathbf{S}_{I},\mathbf{a})
+
\mathcal{L}_{code}(\mathbf{S}_{C},\mathbf{a})
+
\lambda_{hm}\mathcal{L}_{hm},
\end{equation}
where $\mathcal{L}_{code}$ denotes cross-entropy loss for code prediction and $\lambda_{hm}$ balances the heatmap reconstruction term.

\subsection{Using Recovered Heatmaps for FLD}
\label{fldusage}

Following the integration protocol of ORFormer~\cite{orformer}, we use the recovered heatmaps $\mathbf{H}_{rec}$ as additional structural guidance for existing FLD methods~\cite{huang2021adnet,zhou2023star}. Specifically,
$\mathbf{H}_{rec}$ is concatenated with an early-stage feature map of the FLD backbone and fused using the same lightweight convolutional block as ORFormer. The fused representation is then forwarded to the landmark prediction head. We intentionally keep this detector-side fusion unchanged so that the performance gains can be attributed to RCFormer’s heatmap recovery mechanism rather than to a redesigned fusion module.


\section{Experiments}
\label{sec:experiments}

\subsection{Experimental Setup}

\begin{table}[t]
\centering
\small
\setlength{\tabcolsep}{5pt}
\renewcommand{\arraystretch}{1.05}
\begin{tabular}{l l c c c c c c c}
\hline
\multirow{2}{*}{Method} & \multirow{2}{*}{Backbone}
& \multicolumn{3}{c}{WFLW-Full}
& \multicolumn{1}{c}{COFW}
& \multicolumn{3}{c}{300W (NME$\downarrow$)} \\
\cline{3-5}\cline{6-6}\cline{7-9}
& & NME$\downarrow$ & FR$\downarrow$ & AUC$\uparrow$ 
& NME$\downarrow$ & Full & Comm. & Chal. \\
\hline

\multicolumn{9}{c}{\textit{Original WFLW bounding boxes}} \\
\hline

LAB~\cite{wu2018wflw}        & Hourglass  & 5.27 & 7.56 & 0.532 & -    & 3.49 & 2.98 & 5.19 \\
Wing~\cite{feng2018wing}     & ResNet-50  & 4.99 & 6.00 & 0.550 & 5.44 & -    & -    & -    \\
AWing~\cite{wang2019awing}   & Hourglass  & 4.36 & 2.84 & 0.572 & 4.94 & 3.07 & 2.72 & 4.52 \\
PIPNet~\cite{jin2021pipnet}  & ResNet-101 & 4.31 & -    & -     & -    & 3.19 & 2.78 & 4.89 \\
RePFormer~\cite{li2022repformer} & ResNet-101 & 4.11 & - & - & - & 3.01 & - & - \\

\hline
\multicolumn{9}{c}{\textit{HRNet-based WFLW bounding boxes}} \\
\hline

HRNet~\cite{sun2019hrnet}    & HRNet-W18  & 4.60 & 4.64 & -     & -    & 3.32 & 2.87 & 5.15 \\
LUVLi~\cite{kumar2020luvli}  & DU-Net     & 4.37 & 3.12 & 0.577 & -    & 3.23 & 2.76 & 5.16 \\
ADNet~\cite{huang2021adnet}  & Hourglass  & 4.14 & 2.72 & 0.602 & 4.68 & 2.93 & 2.53 & 4.58 \\
HIH~\cite{lan2021hih}        & Hourglass  & 4.08 & 2.60 & 0.605 & 4.63 & 3.09 & 2.65 & 4.89 \\
SLPT~\cite{xia2022slpt}      & HRNet-W18  & 4.14 & 2.76 & 0.595 & 4.79 & 3.17 & 2.75 & 4.90 \\
$\dagger$STAR~\cite{zhou2023star} & Hourglass & 4.03 & 2.32 & 0.611 & 4.62 & \underline{2.90} & \underline{2.52} & 4.46 \\
LDEQ~\cite{micaelli2023ldeq} & Hourglass & 3.92 & 2.48 & \textbf{0.624} & - & - & - & - \\
ORFormer~\cite{orformer} & Hourglass & \underline{3.86} & \underline{1.76} & \underline{0.622} & \underline{4.46} & \underline{2.90} & 2.53 & \underline{4.43} \\
RCFormer & Hourglass & \textbf{3.73} & \textbf{1.65} & \textbf{0.624} & \textbf{4.29} & \textbf{2.81} & \textbf{2.40} & \textbf{4.32} \\

\hline
\end{tabular}

\caption{Comparison with baseline FLD methods. NME is reported for all datasets; FR and AUC at the 10\% threshold are reported for WFLW. WFLW methods are grouped by bounding-box protocol, with direct comparisons and highlighting restricted to the HRNet-based protocol. Best results are shown in bold and second-best results are underlined. $\dagger$ denotes baseline results sourced from ORFormer~\cite{orformer}.}
\label{tab:sota_main}
\end{table}

\paragraph{Datasets and Metrics.}
We evaluate RCFormer on three standard facial landmark detection benchmarks: WFLW~\cite{wu2018wflw}, COFW~\cite{burgos2013cofw}, and 300W~\cite{sagonas2013300w}. WFLW contains 7,500 training images and 2,500 test images annotated with 98 landmarks, and provides subset labels for challenging conditions including pose, expression, illumination, makeup, occlusion, and blur. COFW contains 1,345 training images and 507 test images with 29 landmarks, and is mainly used to evaluate robustness under heavy facial occlusion; although landmark-level occlusion labels are available, we do not use them for training or evaluation. 300W contains 3,148 training images and 689 test images with 68 landmarks, and we follow the standard split of the test set into 554 common and 135 challenging images. Following prior FLD works~\cite{huang2021adnet,zhou2023star}, we report normalized mean error (NME), failure rate (FR), and area under the curve (AUC). Inter-ocular normalization is used for WFLW and 300W, and inter-pupil normalization is used for COFW. The threshold for both FR and AUC is set to 10\%.

\paragraph{Implementation Details.}
The quantized heatmap prior takes a $64\times64\times3$ face image as input and predicts $64\times64\times N_E$ edge heatmaps, where $N_E$ is the number of facial edges. Its latent feature has size $m\times n\times d$, and the codebook contains $N$ entries of dimension $d$. Following ORFormer~\cite{orformer}, we set $m=n=16$, $d=256$, and $N=2048$, and keep the remaining heatmap-prior settings unchanged for fair comparison. RCFormer uses an $L=3$ reliability-aware recovery transformer in this latent space, with token size $1\times1\times d$ and $m\times n$ regular/context tokens. We set $\epsilon=10^{-6}$ in VCC and VGCA. The SRC spatial operator is implemented as a $3\times3$ depthwise convolution to capture local neighborhood information with minimal overhead.

For landmark detection, we follow ADNet~\cite{huang2021adnet} and STAR~\cite{zhou2023star}, using a four-stacked hourglass network~\cite{newell2016stacked} as the backbone. The recovered heatmaps are concatenated with the backbone feature maps and fused by a lightweight $1\times1$ convolutional block before the first hourglass module. All models are trained from scratch on the target dataset without external data or pretrained weights.


\begin{figure*}[t]
\centering
\includegraphics[width=0.65\linewidth]{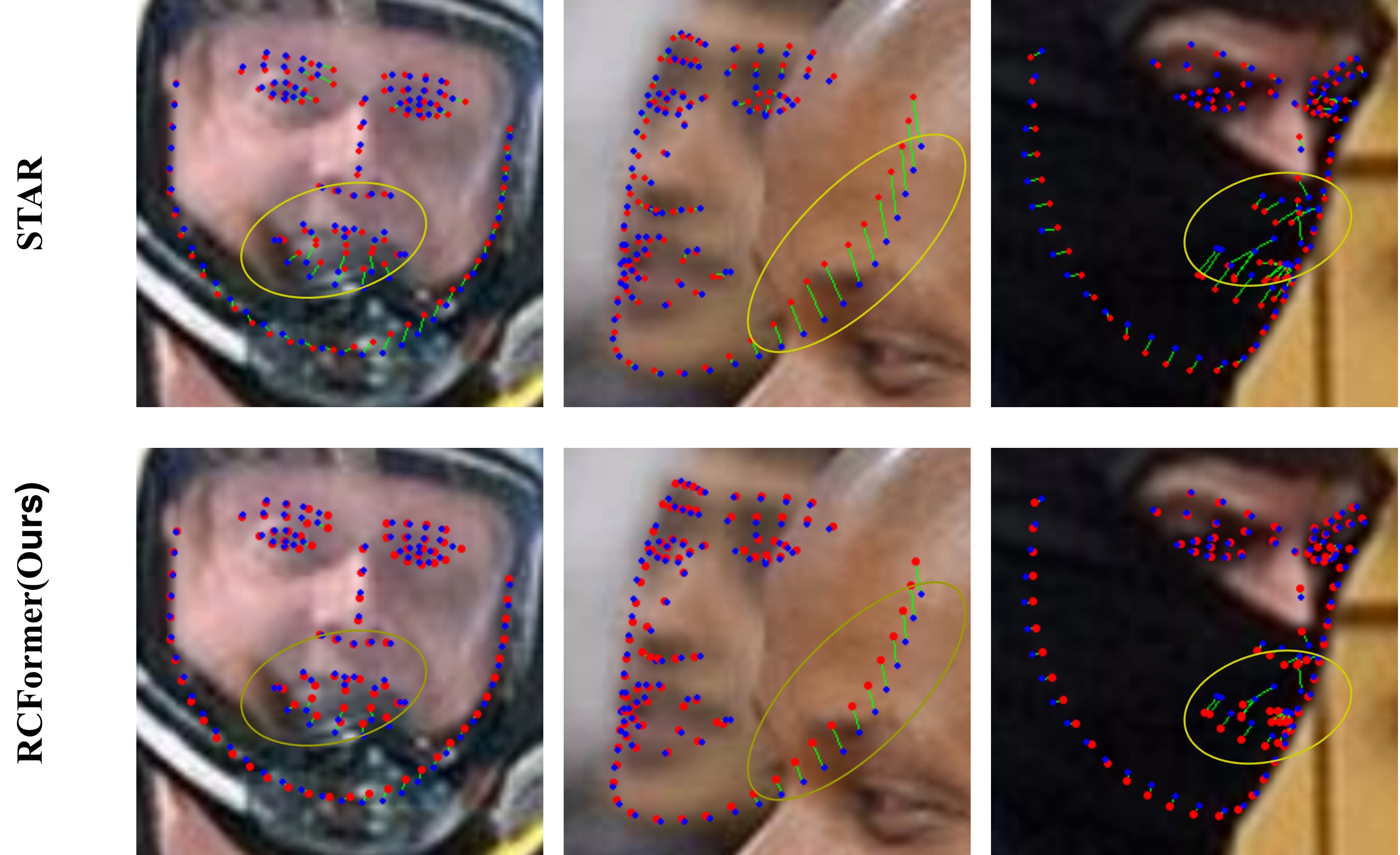}
\caption{
Qualitative landmark comparison on challenging WFLW samples. Blue points denote ground-truth landmarks, red points denote predicted landmarks, and green line segments show the displacement between matched landmark pairs. Yellow ellipses highlight difficult regions where RCFormer produces shorter landmark-error lines than STAR.}
\label{fig:qual_landmark}
\end{figure*}
\begin{table}[t]
\small
\centering
\setlength{\tabcolsep}{5pt}
\renewcommand{\arraystretch}{1.05}
\begin{tabular}{l l c c c c c c c}
\hline
\multirow{2}{*}{\textbf{Method}} & \multirow{2}{*}{\textbf{Backbone}}
& \textbf{Full} & \textbf{Pose} & \textbf{Exp.} & \textbf{Ill.} & \textbf{Mak.} & \textbf{Occ.} & \textbf{Blur} \\
\cline{3-9}
& & \multicolumn{7}{c}{\textbf{NME}$\downarrow$} \\
\hline
\multicolumn{9}{c}{\textit{Original WFLW bounding boxes}} \\
\hline
LAB~\cite{wu2018wflw}        & Hourglass  & 5.27 & 10.24 & 5.51 & 5.23 & 5.15 & 6.79 & 6.12 \\
Wing~\cite{feng2018wing}     & ResNet-50  & 4.99 & 8.43  & 5.21 & 4.88 & 5.26 & 6.21 & 5.81 \\
AWing~\cite{wang2019awing}   & Hourglass  & 4.36 & 7.38  & 4.58 & 4.32 & 4.27 & 5.19 & 4.96 \\
PIPNet~\cite{jin2021pipnet}  & ResNet-101 & 4.31 & 7.51  & 4.44 & 4.19 & 4.02 & 5.36 & 5.02 \\
RePFormer~\cite{li2022repformer} & ResNet-101 & 4.11 & 7.25 & 4.22 & 4.04 & 3.91 & 5.11 & 4.76 \\

\hline
\multicolumn{9}{c}{\textit{HRNet-based WFLW bounding boxes}} \\
\hline
HRNet~\cite{sun2019hrnet}     & HRNet-W18  & 4.60 & 7.94 & 4.85 & 4.55 & 4.29 & 5.44 & 5.42 \\
ADNet~\cite{huang2021adnet}   & Hourglass  & 4.14 & 6.96 & 4.38 & 4.09 & 4.05 & 5.06 & 4.79 \\
HIH~\cite{lan2021hih}         & Hourglass  & 4.08 & 6.87 & 4.06 & 4.34 & 3.85 & 4.85 & 4.66 \\
SLPT~\cite{xia2022slpt}       & HRNet-W18  & 4.14 & 6.96 & 4.45 & 4.05 & 4.00 & 5.06 & 4.79 \\
$\dagger$STAR~\cite{zhou2023star} & Hourglass & 4.03 & 6.78 & 4.26 & 3.97 & 3.85 & 4.82 & 4.59 \\
LDEQ~\cite{micaelli2023ldeq}  & Hourglass  & 3.92 & 6.86 & \textbf{3.94} & 4.17 & 3.75 & 4.77 & 4.59 \\
ORFormer~\cite{orformer}      & Hourglass  & \underline{3.86} & \underline{6.63} & 4.13 & \underline{3.83} & \underline{3.61} & \underline{4.57} & \underline{4.50} \\
RCFormer (Ours)               & Hourglass  & \textbf{3.73} & \textbf{6.48} & \underline{3.96} & \textbf{3.71} & \textbf{3.49} & \textbf{4.38} & \textbf{4.36} \\
\hline
\end{tabular}
\caption{
WFLW subset evaluation. NME is reported on the full test set and six challenging subsets. Methods are grouped by bounding-box protocol, with direct comparisons and highlighting restricted to the HRNet-based protocol. Best results are shown in bold and second-best results are underlined. $\dagger$ denotes baseline results sourced from ORFormer~\cite{orformer}.}
\label{tab:wflw_subsets}
\end{table}

\begin{table}[t]
\centering
\small
\setlength{\tabcolsep}{3pt}
\renewcommand{\arraystretch}{1.02}
\begin{tabular}{l c c c c c c c c c}
\toprule
\multirow{2}{*}{Architecture}
& \multicolumn{7}{c}{WFLW}
& COFW
& 300W \\
\cmidrule(lr){2-8}\cmidrule(lr){9-9}\cmidrule(lr){10-10}
& Full & Pose & Exp. & Ill. & Mak. & Occ. & Blur & Full & Full \\
\midrule
\multicolumn{10}{c}{Heatmap L2 Loss $\downarrow$} \\
\midrule
VQVAE    & 29.38 & 59.37 & 32.64 & 28.45 & 33.25 & 40.12 & 34.23 & 16.57 & 25.97 \\
ORFormer & 22.29 & 43.25 & 26.49 & 22.07 & 23.31 & 28.48 & 27.05 & 12.80 & 16.58 \\
RCFormer & 15.22 & 25.41 & 18.46 & 16.56 & 16.75 & 19.12 & 17.20 & 8.39  & 12.59 \\
\midrule
\footnotesize\textbf{Rel. gain vs. ORFormer}
& \footnotesize\textbf{31.7\%}
& \footnotesize\textbf{41.2\%}
& \footnotesize\textbf{30.3\%}
& \footnotesize\textbf{25.0\%}
& \footnotesize\textbf{28.1\%}
& \footnotesize\textbf{32.9\%}
& \footnotesize\textbf{36.4\%}
& \footnotesize\textbf{34.5\%}
& \footnotesize\textbf{24.1\%} \\
\bottomrule
\end{tabular}
\caption{
Heatmap generation comparison using L2 loss. All architectures are implemented under the same setting. Results are reported on WFLW and its six subsets, with full-set results on COFW and 300W. Bold percentages indicate relative L2 reduction of RCFormer over ORFormer.
}
\label{tab:ablate_l2loss}
\end{table}
Following ORFormer~\cite{orformer}, we use Adam~\cite{kingma2014adam} with a cosine annealing warm-restart scheduler~\cite{sgdr2017} for all stages, using the same restart period of 5 and period multiplier of 2. Training is performed on a single NVIDIA RTX 4090 GPU with 24GB memory, and all stages are trained for 300 epochs. For the quantized heatmap prior, the batch size is 128 for all datasets. The learning rate is $7\times10^{-4}$ for WFLW and COFW, and $8\times10^{-4}$ for 300W. We set $\beta=0.25$ for all datasets, with $\lambda_{vq}=100$ for WFLW and COFW and $\lambda_{vq}=1000$ for 300W. For RCFormer, the batch size is 64 and the learning rate is $1\times10^{-4}$ for all datasets. We set $\lambda_{hm}=50$ for WFLW and COFW, and $\lambda_{hm}=100$ for 300W. For the downstream FLD detector, the batch size is 16 and the learning rate is $1\times10^{-3}$ for all datasets. 


\paragraph{Data Augmentation.} Following common training protocols in facial landmark detection and face recovery~\cite{huang2021adnet,zhou2023star,codeformer}, we crop the face region and resize it to $64\times64$ for the heatmap-prior and RCFormer stages. We apply random horizontal flipping with probability $0.5$, random grayscale conversion with probability $0.5$, rotation within $\pm30^\circ$, translation within $\pm4\%$, and scaling within $\pm5\%$. Following~\cite{orformer}, an additional random occlusion is inserted into each image during RCFormer training to simulate unreliable facial regions; these synthetic occlusions are used only as augmentation. For the downstream landmark detector, images are resized to $256\times256$, and random occlusion is applied with probability $0.4$ while keeping the remaining augmentations unchanged. Improvements on the WFLW occlusion subset and COFW, both containing real occlusions, indicate that RCFormer does not merely learn synthetic occlusion patterns.

\subsection{Comparison with Existing Methods}
Table~\ref{tab:sota_main} reports the comparison on WFLW, COFW, and 300W. For WFLW, we separate methods according to the bounding-box protocol and restrict direct comparisons to methods evaluated using HRNet-based boxes. Under this protocol, RCFormer achieves the lowest NME and FR while matching the best AUC. RCFormer also obtains the lowest NME on COFW and all 300W splits. These results demonstrate consistent improvements across the evaluated FLD benchmarks.

For qualitative evaluation, we visualize landmark predictions on challenging WFLW samples in Figure~\ref{fig:qual_landmark}. We compare RCFormer with STAR~\cite{zhou2023star}, which uses the same hourglass backbone. In the highlighted regions, RCFormer generally shows shorter landmark-error lines than STAR, particularly around occluded or weakly visible facial boundaries. This suggests that the recovered heatmaps provide useful structural guidance for landmark prediction in challenging cases.

\begin{figure*}[t]
\centering
\includegraphics[width=0.6\linewidth]{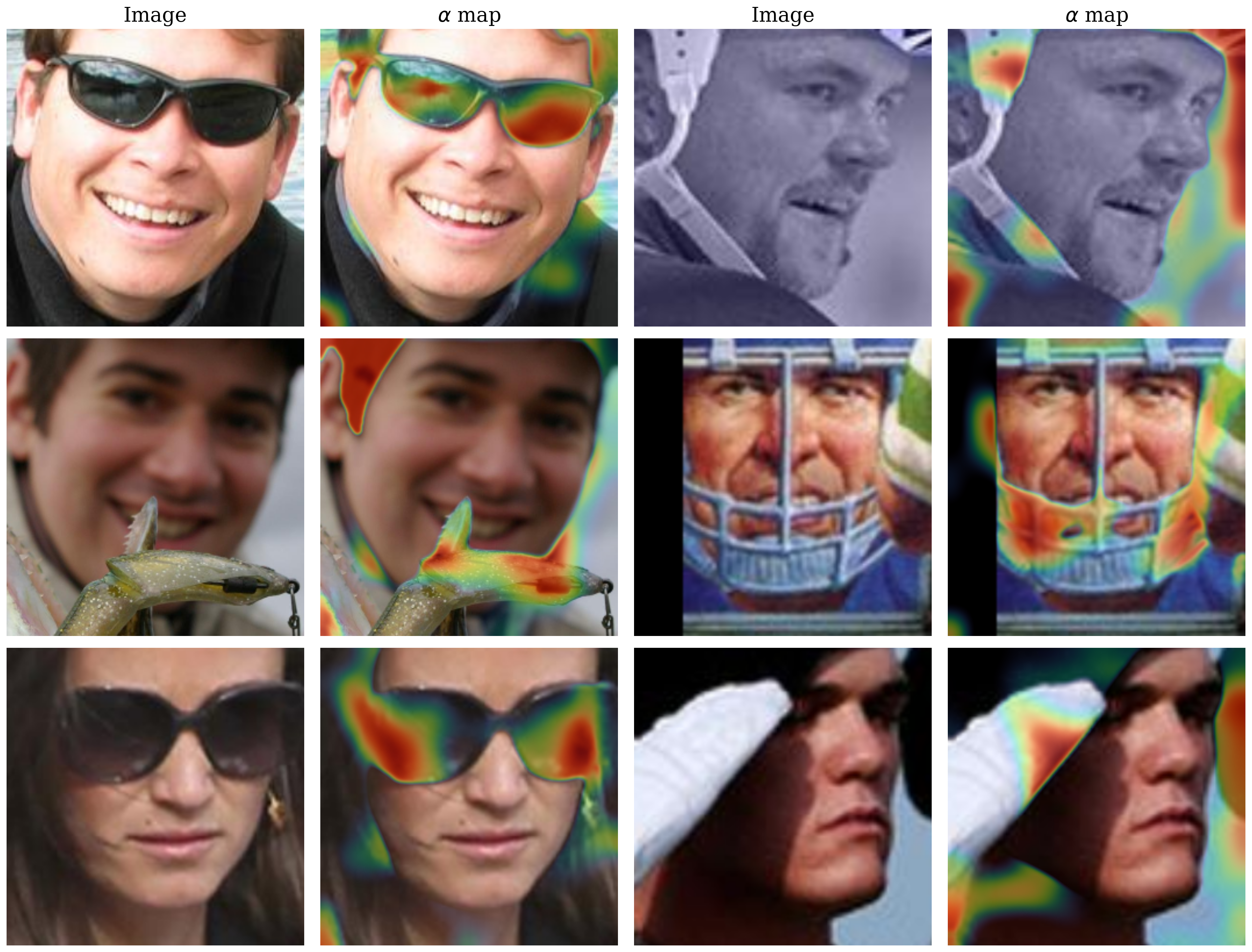}
\caption{
Predicted unreliability maps from RCFormer. Red regions indicate higher unreliability, highlighting occlusions, low-visibility areas, and ambiguous facial boundaries.
}
\label{fig:alpha_maps}
\end{figure*}

Furthermore, we evaluated RCFormer on the challenging subsets of WFLW in Table~\ref{tab:wflw_subsets}. RCFormer improves over ORFormer on the full test set and achieves lower NME on all six subsets, showing that the proposed recovery design is effective across different challenging conditions. On the expression subset, RCFormer still improves over ORFormer but ranks second overall. This may indicate that expression variation is less directly addressed by visibility-oriented recovery than occlusion, pose, illumination, makeup, or blur. We also visualize the predicted unreliability maps in Figure~\ref{fig:alpha_maps}. High-response regions tend to appear around occlusions, low-visibility areas, and ambiguous facial boundaries, while more reliable facial regions show lower responses. This suggests that the learned unreliability maps correlate with visually degraded regions and provide spatial guidance for feature recovery.



\begin{figure}[t]
\centering
\includegraphics[width=1 \linewidth]{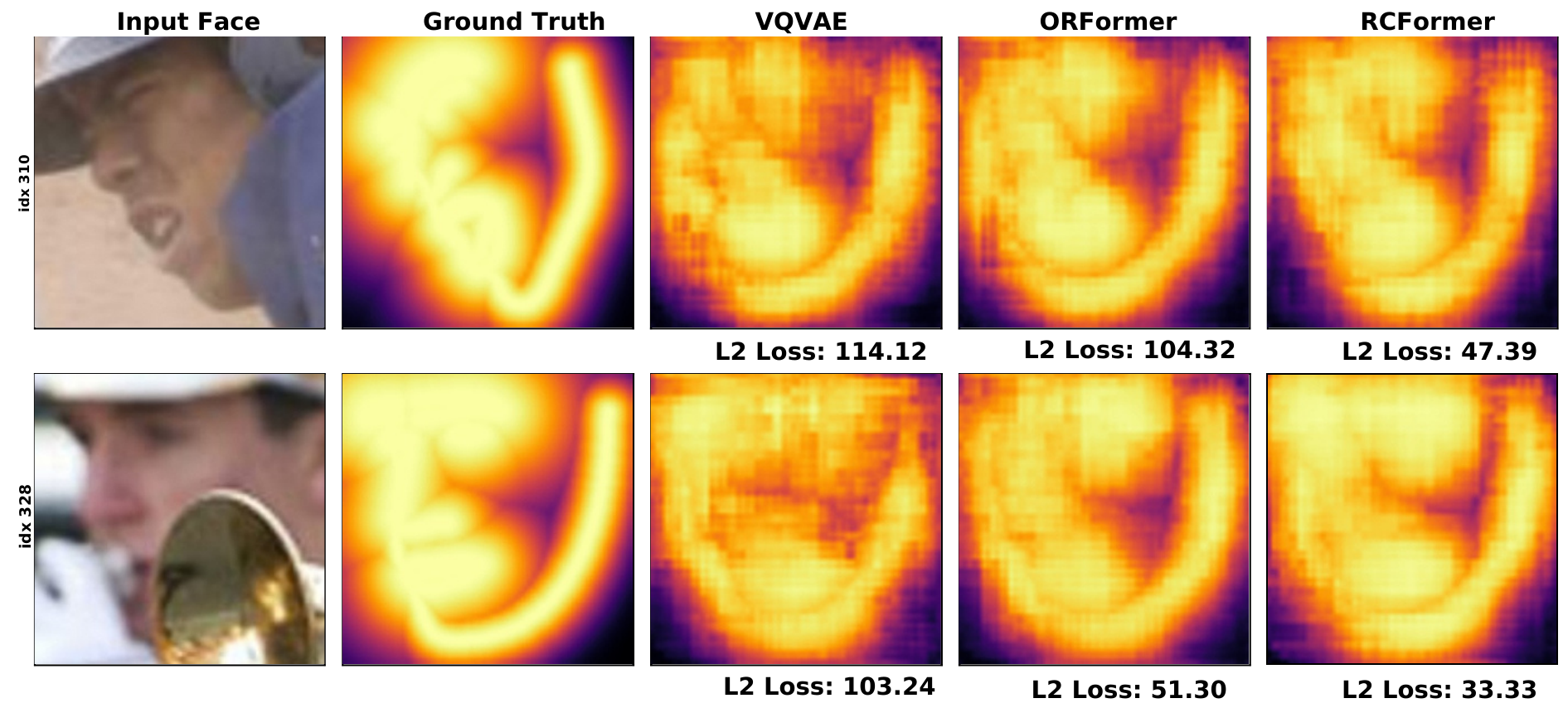}
\caption{Edge heatmap outputs on WFLW samples. Ground truth (GT) heatmaps are derived from landmark-based facial edge annotations, with brighter regions indicating stronger responses. Values denote per-sample L2 errors against GT.}

\label{fig:heatmap_comparison}
\end{figure}

\subsection{Ablation Studies}



\paragraph{Recovered Heatmap Quality.}

To evaluate heatmap generation quality, we report the heatmap L2 loss in Table~\ref{tab:ablate_l2loss}. The comparison includes VQVAE as the quantized heatmap generator baseline and ORFormer as the context-guided latent recovery baseline. ORFormer improves over VQVAE, confirming the benefit of context-guided latent patch recovery. RCFormer further reduces the L2 loss from 22.29 to 15.22 on WFLW-Full and achieves relative gains from 24.1\% to 41.2\% over ORFormer across the evaluated settings. These reductions show that reliability-aware context recovery substantially improves recovered heatmap quality. The gains are larger for heatmap recovery than for NME because RCFormer directly optimizes the recovered heatmaps, while the downstream detector uses them only as structural guidance and remains constrained by detector capacity and annotation ambiguity. Nevertheless, the consistent NME improvements in Tables~\ref{tab:sota_main} and~\ref{tab:wflw_subsets} confirm their downstream benefit. Figure~\ref{fig:heatmap_comparison} directly compares the recovered heatmaps from VQVAE, ORFormer, and RCFormer with ground truth. RCFormer produces cleaner facial structures and lower per-sample L2 errors than the baselines.

\begin{table}[t]

\centering
\small
\setlength{\tabcolsep}{5pt}
\renewcommand{\arraystretch}{1.05}
\begin{tabular}{l c c c c c}
\toprule
\multirow{2}{*}{Method}
& \multicolumn{3}{c}{Components}
& \multicolumn{2}{c}{WFLW} \\
\cmidrule(lr){2-4}\cmidrule(lr){5-6}
& VCC & VGCA & SRC & L2 Loss$\downarrow$ & NME$\downarrow$ \\
\midrule
Base             & -- & -- & -- & 22.05 & 3.96 \\
Base + VCC       & \checkmark & -- & -- & 21.08 & 3.90 \\
Base + VCC + VGCA & \checkmark & \checkmark & -- & 20.14 & 3.86 \\
Base + SRC       & -- & -- & \checkmark & 18.46 & 3.89 \\
RCFormer         & \checkmark & \checkmark & \checkmark & 15.22 & 3.73 \\
\bottomrule
\end{tabular}
\caption{
Component ablation on WFLW. VCC, VGCA, and SRC denote visible-context conditioning, visibility-guided context aggregation, and spatial residual correction, respectively.
}
\label{tab:component_ablation}
\end{table}

\paragraph{Component Analysis of RCFormer.}
We evaluate the contribution of each RCFormer component on WFLW using heatmap L2 loss and landmark NME. The base model keeps the regular/context token streams, hard self-exclusion context attention in Eq.~\ref{eq:self_excluded_context_attention}, unreliability estimation, dual code prediction, and patch-wise latent blending, but removes VCC, VGCA, and SRC. As shown in Table~\ref{tab:component_ablation}, adding VCC improves both metrics, showing that conditioning the context query with a visibility-weighted global context makes the recovered context more adaptive to the input face. Adding VGCA on top of VCC further improves both metrics, suggesting that visibility-guided context aggregation complements visible-context conditioning. 
SRC gives a strong standalone reduction in heatmap L2 loss by applying spatial correction under the guidance of the unreliability map, although its NME gain is limited when using the base reliability estimates. Combining SRC with VCC and VGCA gives the best L2 loss and NME, suggesting that reliability-aware context formation and spatial correction work together in RCFormer.



\begin{table}[t]
\centering
\small
\setlength{\tabcolsep}{7pt}
\renewcommand{\arraystretch}{1.05}

\begin{tabular}{lcccc}
\toprule
Method & Params (M) & FLOPs (G) & Latency (ms) & NME$\downarrow$ \\
\midrule
STAR     & 17.177 & 34.796 & 34.154 & 4.03 \\
ORFormer & 22.038 & 36.260 & 37.033 & 3.86 \\
RCFormer & 22.699 & 36.264 & 37.931 & \textbf{3.73} \\
\bottomrule
\end{tabular}

\caption{Inference efficiency on WFLW measured on an NVIDIA RTX 4090 with batch size 1 and FP32 precision.}
\label{tab:inference_efficiency}
\end{table}

\paragraph{Computational Efficiency.}
Table~\ref{tab:inference_efficiency} reports the complete inference cost of STAR, ORFormer, and RCFormer on WFLW. Compared with ORFormer, RCFormer adds only 0.004 G FLOPs and 0.898 ms latency while reducing NME from 3.86 to 3.73; peak GPU memory remains similar at 145.25 MB versus 143.48 MB. Compared with STAR, RCFormer adds 1.468 G FLOPs and 3.777 ms latency. At the recovery-network level, RCFormer has 5.31M parameters compared with 4.78M for ORFormer, an increase of approximately 0.53M (11.0\%), with VCC and SRC accounting for the additional parameters while VGCA introduces no learnable parameters. Overall, these results show that the improved heatmap recovery and landmark localization are obtained with modest additional computational cost.

\section{Conclusion, Limitations and Future Work}

We introduced RCFormer, a reliability-aware extension of context-guided latent patch recovery for occlusion-robust FLD. By combining visible-context conditioning, visibility-guided context aggregation, and spatial residual correction, RCFormer improves both recovered heatmap quality and landmark localization. Experiments on WFLW, COFW, and 300W demonstrate consistent improvements over strong baselines across challenging facial conditions, supporting the effectiveness of reliability-aware context recovery. The recovery mechanism may become less effective when occlusions are extremely large, reliable contextual evidence is scarce, or pose and expression variations dominate the localization error. While the results across three benchmarks support the robustness of RCFormer, extending the evaluation to cross-dataset settings would provide further insight into its generalization under unseen data distributions and represents a natural direction for future work.



\bibliography{egbib}


\end{document}


\maketitle

\section{Supplementary Materials}

\subsection{Heatmap Definition}

RCFormer aims to identify unreliable facial regions and recover missing structural information from degraded face images, enabling the generation of high-fidelity heatmaps under challenging conditions such as occlusion, extreme pose, blur, and lighting variation. These recovered heatmaps provide facial contour guidance that helps facial landmark detection (FLD) methods maintain robustness in challenging scenarios.

Following ORFormer~\cite{orformer}, which adopts the edge-based heatmap representation of Wu et al.~\cite{wu2018wflw}, we use the same heatmap construction protocol and dataset-specific landmark-to-edge mappings. Using edge heatmaps reduces computational cost while preserving sufficient structural information for downstream FLD methods.

\subsubsection{Heatmap Generation}

For a given face image $I \in \mathbb{R}^{h \times w \times 3}$ and its ground-truth landmark annotations
$L = \{l_i\}_{i=0}^{N_L-1}$, we divide $L$ into $N_E$ subsets
$L_j \subset L$, where $j = 0, \ldots, N_E - 1$, to represent facial edges such as the cheek, eyebrow, eye contour, nose, and mouth contour. Here, $N_L$ denotes the number of landmarks per face, and $N_E$ denotes the number of facial edges.

Each facial edge $L_j$ is used to interpolate an edge line, forming a binary edge map $B_j$ with the same spatial size as the face image. We then apply a distance transform to $B_j$, which computes the nearest distance from each pixel to the corresponding edge line and produces the distance map $M_j$. The ground-truth edge heatmap $\hat{H}_j$ used to supervise the quantized heatmap generator and RCFormer is defined as
\begin{equation}
\hat{H}_j(x,y)=
\begin{cases}
\exp\left(-\dfrac{M_j(x,y)^2}{2\sigma^2}\right), & \text{if } M_j(x,y) < 3\sigma,\\[6pt]
0, & \text{otherwise}.
\end{cases}
\label{eq:edge_heatmap}
\end{equation}
where $\sigma$ denotes the standard deviation of the values in the distance map $M_j$.

\subsubsection{Index Mapping}

Our experiments are conducted on three datasets: WFLW~\cite{wu2018wflw}, COFW~\cite{burgos2013cofw}, and 300W~\cite{sagonas2013300w}. Since the number of landmarks varies across these datasets, the facial edge heatmap definitions are dataset-specific. Therefore, we provide the landmark-to-edge index mappings below.

For the WFLW dataset, with $N_L=98$ and $N_E=15$, the index mapping is given as follows:
\begin{align*}
\text{Edge }0 &: [0\text{-}32] \\
\text{Edge }1 &: [33\text{-}37] \\
\text{Edge }2 &: [38\text{-}41,33] \\
\text{Edge }3 &: [42\text{-}46] \\
\text{Edge }4 &: [46\text{-}49,50] \\
\text{Edge }5 &: [51\text{-}54] \\
\text{Edge }6 &: [55\text{-}59] \\
\text{Edge }7 &: [60\text{-}64] \\
\text{Edge }8 &: [64\text{-}67,60] \\
\text{Edge }9 &: [68\text{-}72] \\
\text{Edge }10 &: [72\text{-}75,68] \\
\text{Edge }11 &: [76\text{-}82] \\
\text{Edge }12 &: [82\text{-}87,76] \\
\text{Edge }13 &: [88\text{-}92] \\
\text{Edge }14 &: [92\text{-}95,88].
\end{align*}

For the 300W dataset, with $N_L = 68$ and $N_E = 13$, the index mapping is
\begin{align*}
\text{Edge }0 &: [0\text{-}16] \\
\text{Edge }1 &: [17\text{-}21] \\
\text{Edge }2 &: [22\text{-}26] \\
\text{Edge }3 &: [27\text{-}30] \\
\text{Edge }4 &: [31\text{-}35] \\
\text{Edge }5 &: [36\text{-}39] \\
\text{Edge }6 &: [39\text{-}41,36] \\
\text{Edge }7 &: [42\text{-}45] \\
\text{Edge }8 &: [45\text{-}47,42] \\
\text{Edge }9 &: [48\text{-}54] \\
\text{Edge }10 &: [54\text{-}59,48] \\
\text{Edge }11 &: [60\text{-}64] \\
\text{Edge }12 &: [64\text{-}67,60].
\end{align*}

For the COFW dataset, with $N_L = 29$ and $N_E = 14$, the index mapping is
\begin{align*}
\text{Edge }0 &: [0,4,2] \\
\text{Edge }1 &: [2,5,0] \\
\text{Edge }2 &: [1,6,3] \\
\text{Edge }3 &: [3,7,1] \\
\text{Edge }4 &: [8,12,10] \\
\text{Edge }5 &: [10,13,8] \\
\text{Edge }6 &: [9,14,11] \\
\text{Edge }7 &: [11,15,9] \\
\text{Edge }8 &: [18,21,19] \\
\text{Edge }9 &: [20,21] \\
\text{Edge }10 &: [22,26,23] \\
\text{Edge }11 &: [23,27,22] \\
\text{Edge }12 &: [22,24,23] \\
\text{Edge }13 &: [23,25,22].
\end{align*}

\bibliography{egbib}